\documentclass[letterpaper, 10 pt, conference]{ieeeconf}  

\IEEEoverridecommandlockouts                              

\usepackage{graphicx}
\usepackage{amsmath}
\usepackage{amssymb}
\usepackage{booktabs}
\usepackage{cite}
\usepackage[dvipsnames]{xcolor}
\usepackage{multirow}
\usepackage{hhline}
\usepackage{hyperref}
\hypersetup{
    colorlinks=true,
    linkcolor=black,
    filecolor=magenta,      
    urlcolor=black}
\usepackage[capitalize]{cleveref}
\usepackage{subcaption}
\title{\LARGE \bf
Viewer-Centred Surface Completion for Unsupervised Domain Adaptation in 3D Object Detection
}

\author{Darren Tsai, Julie Stephany Berrio, Mao Shan, Eduardo Nebot and Stewart Worrall%
\thanks{This work has been supported by the Australian Centre for Field Robotics (ACFR), Baraja Pty Ltd. and ARC LIEF grant LE200100049 Whopping Volta GPU Cluster - Transforming Artificial Intelligence Research. (Corresponding author: Darren Tsai.)}
\thanks{The authors are with the Australian Centre for Field Robotics (ACFR) at the University of Sydney (NSW, Australia). E-mails: {\small{\{d.tsai, j.berrio, m.shan, e.nebot, s.worrall}\}@acfr.usyd.edu.au}}%
}

\begin{document}

\maketitle
\thispagestyle{empty}
\pagestyle{empty}

\begin{abstract}

Every autonomous driving dataset has a different configuration of sensors, originating from distinct geographic regions and covering various scenarios. As a result, 3D detectors tend to overfit the datasets they are trained on. This causes a drastic decrease in accuracy when the detectors are trained on one dataset and tested on another. We observe that lidar scan pattern differences form a large component of this reduction in performance. We address this in our approach, SEE-VCN, by designing a novel viewer-centred surface completion network (VCN) to complete the surfaces of objects of interest within an unsupervised domain adaptation framework, SEE \cite{tsai2022see}. With SEE-VCN, we obtain a unified representation of objects across datasets, allowing the network to focus on learning geometry, rather than overfitting on scan patterns. By adopting a domain-invariant representation, SEE-VCN can be classed as a multi-target domain adaptation approach where no annotations or re-training is required to obtain 3D detections for new scan patterns. Through extensive experiments, we show that our approach outperforms previous domain adaptation methods in multiple domain adaptation settings. Our code and data are available at \url{https://github.com/darrenjkt/SEE-VCN}.

\end{abstract}

\section{Introduction}

Several large-scale autonomous driving datasets have been released by companies \cite{sun2020waymo,geiger2012kitti,caesar2020nuscenes,houston2020lyft}. Each has a different sensor configuration, collected in distinct geographic regions with unique scenarios. The differences in each of these dataset-specific factors leads to a degradation in performance for 3D detectors when training on one dataset and testing on another. Many works have been proposed to tackle different aspects of this domain gap such as addressing lidar scan pattern discrepancies \cite{tsai2022see,wei2022lidardistillation}, object size \cite{wang2020train}, range association \cite{zhang2021srdan} and weather \cite{xu2021spg}. These unsupervised domain adaptation (UDA) approaches seek to bridge the domain gap between datasets by adapting a model trained on labelled point clouds (source domain) to unlabelled point clouds (target domain).

Unlike images which are structured in a 2D grid representation, point cloud data is unstructured when representing objects in 3D space. Each lidar has their own specific vertical/horizontal angular resolution, number of beams, methods for handling of artefacts, noise profile and range capabilities. This scan pattern domain gap across lidars leads to a representation of objects that is unique for each lidar, confusing 3D detectors in identifying similar objects.

With the ongoing innovation of lidars \cite{baraja2020next,lambert2020lidarcomparison}, newer robots or vehicles will be configured with lidars that have different scan patterns. Each new lidar would require substantial time and expenses for annotation and re-training. A unique kind of lidar explored in our previous work \cite{tsai2022see} is the Baraja Spectrum-Scan™, which can adjust its scan pattern in real-time for user-defined vertical and horizontal angular resolution. With this lidar, we require a model that can adapt to any user-defined scan pattern. Needing to dispatch specific models for each scan pattern limits the flexibility of such a lidar. We build upon this work to have a single model that can perform well on any scan pattern, without requiring labels or re-training for unseen scan patterns.

\begin{figure}[t]
  \centering
   \includegraphics[width=0.96\linewidth]{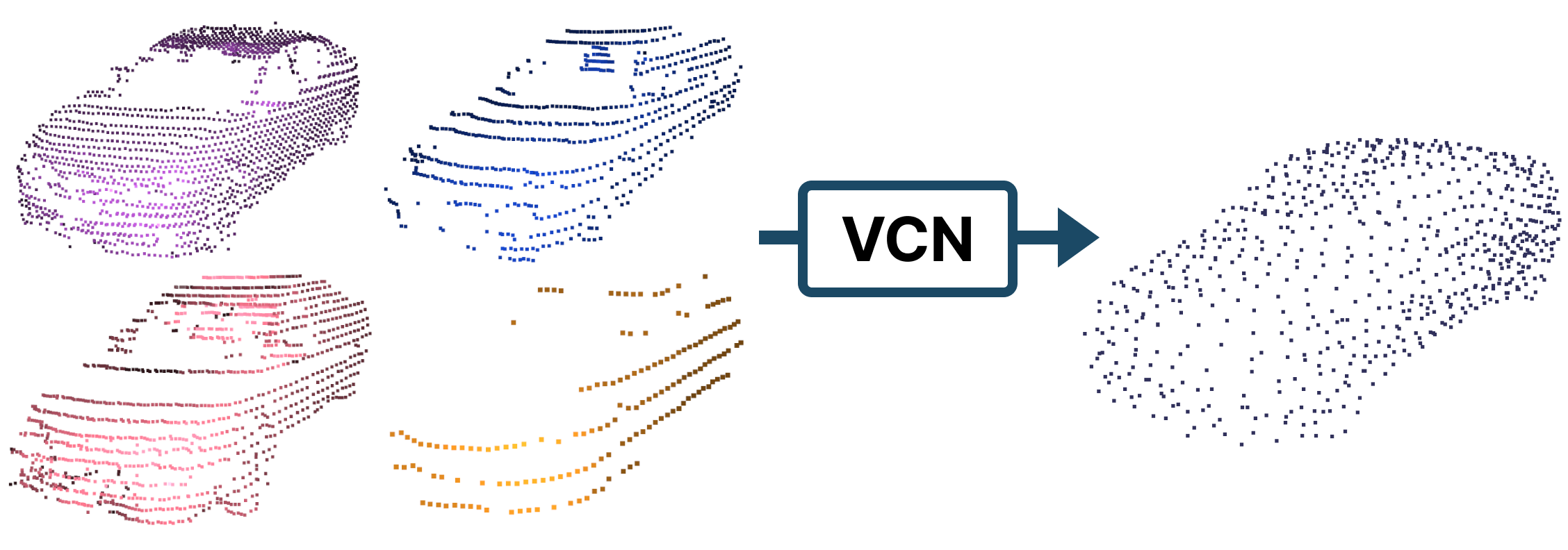}
   \caption{VCN transforms objects from each lidar, into a unified surface representation.}
   \label{fig:vcn_illustration}
   \vspace{-4mm}
\end{figure}

Multiple previous works have proposed to transform the point cloud from the source (training) domain to match the point cloud scan pattern of the target (testing) domain. A naive approach to this strategy is to randomly downsample the point cloud or skip lidar beams when adapting from a high to low resolution lidar. However, this has been observed to be an unsatisfactory representation  \cite{xu2021spg} of the low resolution lidar scan pattern. \cite{wei2022lidardistillation} proposes a knowledge distillation \cite{hinton2015distilling} framework to generate low-beam pseudo-point clouds by downsampling the high-beam source point clouds. Nonetheless, this method does not address the alternate scenario of low to high beam nor the difference in angular resolution across same-beam lidars. In the low to high resolution lidar scenario, the point cloud can be upsampled \cite{li2019pu,yifan2019patch}, however these networks struggle with recovering the 3D geometry of partially observed objects. Furthermore, upsampling the entire point cloud also leads to higher latency. Instead of augmenting the entire point cloud, we propose an approach that selectively augments parts of the point cloud to obtain good performance and reduce pre-processing latency.

We focus on addressing this scan pattern discrepancy aspect of the domain gap with our proposed approach, SEE-VCN. SEE-VCN enables consistent representation of objects across lidars for multi-target UDA in 3D object detection. We achieve this by incorporating the SEE framework \cite{tsai2022see} with our lidar-agnostic, viewer-centred surface completion network (VCN). VCN extrapolates the underlying surface of each point to transform the lidar-specific to a unified representation, shown in \cref{fig:vcn_illustration}. 

Deep learning point cloud completion approaches \cite{yuan2018pcn,yu2021pointr,groueix2018atlas,yang2018foldingnet} are typically applied in object-centred coordinates, where ground truth orientation and dimensions are required to pre-canonicalize objects to a unit-bounding box with a pose that is centred and aligned to the Euclidean frame axes. VCN on the other hand, can estimate the pose and complete the surface of an object without pre-canonicalization. Overall, SEE-VCN outperforms multi-target domain adaptation (MTDA) approach, SEE, and single-target domain adaptation (STDA) approach ST3D \cite{yang2021st3d} in multiple domain adaptation settings.


\begin{figure*}
  \vspace{2mm}
  \centering
  \includegraphics[width=0.99\linewidth]{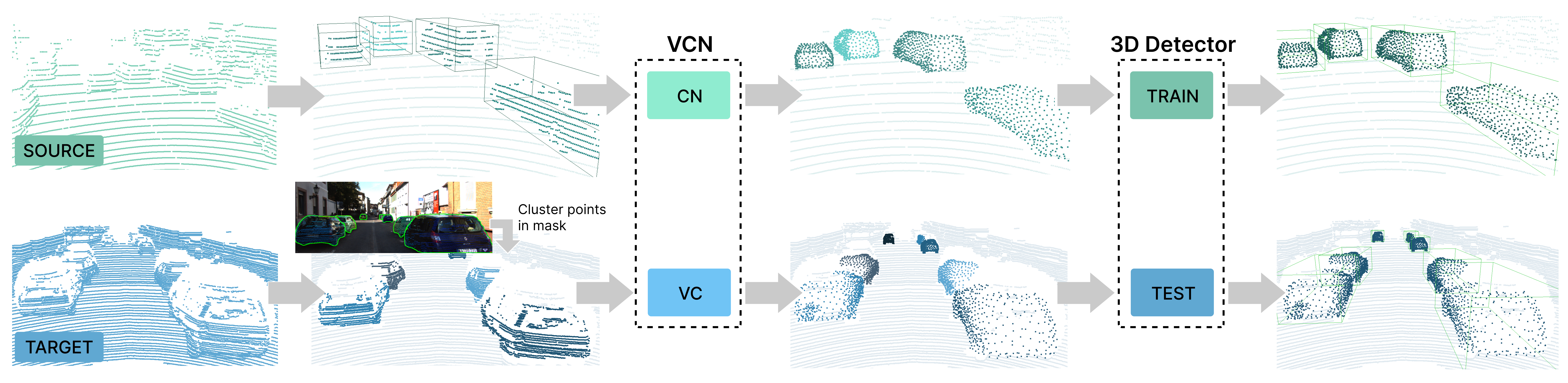}
  \caption{SEE-VCN integrates the SEE framework with our Viewer-centred Completion Network (VCN) to transform objects of interest to a unified point cloud representation. For the source domain, we use ground truth labels to transform objects into an object-centred (canonical) coordinates before completion (VCN-CN). For the target domain, we estimate the object pose to transform the viewer-centred object to object-centred coordinates before completing the object (VCN-VC).}
  \label{fig:framework}
  \vspace{-4mm}
\end{figure*}

\section{Related Works}
\subsection{Unsupervised Domain Adaptation}
The goal of UDA is to adapt a model trained on a labelled source domain to generalize well on an unlabelled target domain. There are a few variations of this UDA task. Single-target DA \cite{yang2021st3d, zhang2021srdan, wang2020train,wu2019squeezesegv2} adapts only one labelled source dataset to one unlabelled target dataset. Multi-source DA \cite{zhao2018adversarial} uses multiple labelled source datasets, to adapt to one target dataset. Source-free DA \cite{saltori2020sf} adapts an already trained model on the source domain, to an unlabelled target domain without using any source labels. Multi-target DA \cite{yi2021complete,tsai2022see} adapts one labelled source domain, to multiple unlabelled target domains. Across these different UDA settings, there are several common strategies. Domain-invariant feature representation is a common approach where explicit distance measures such as maximum mean discrepancy \cite{rozantsev2018beyond} and correlation alignment \cite{rozantsev2018beyond} are used to learn feature representations that are invariant between the two domains. Domain-invariant features can also be learnt with adversarial training \cite{ganin2015unsupervised,saito2018maximum,qin2019generatively,sankaranarayanan2018generate}. Instead of aligning the feature space, some methods align the data representation. This could be in the form of using adversarial networks to align the appearance of the source to resemble the target domain \cite{isobe2021multi,hong2018conditional}, or in transforming all input data into a domain-invariant representation \cite{yi2021complete,tsai2022see,li2019pu}. Another category of works explore self-training \cite{yang2021st3d,zou2018unsupervised,lian2019constructing,li2020content} to generate pseudo-labels for training on the target dataset and is popular in 3D object detection STDA approaches \cite{yang2021st3d,yang2021st3d++,luo2021unsupervised,ding2022jst,you2022exploiting}. \cite{you2022exploiting,fruhwirth2021fast3d} leveraged object tracking across multiple frames to generate higher quality pseudo-labels for self-training. 

The MTDA task is a more challenging setting than STDA, as it is difficult to find features that generalize well to multiple target domains.  \cite{isobe2021multi,nguyen2021unsupervised} propose to distill domain knowledge from multiple experts (teachers) to a shared student for image segmentation. For point cloud tasks, a few works propose to transform the entire input point cloud into a domain-invariant representation by completing its surface \cite{yi2021complete,li2019pu}. However, transforming the entire point cloud is computationally expensive. In our previous work, we proposed a partial point cloud transformation by leveraging image detectors to transform objects into a domain-invariant representation selectively. However, the ball pivoting algorithm used in the work is slow and leads to inconsistent car geometries. Instead, VCN is fast (0.32ms/car) and generates consistent surface representations for cars from different lidars. Overall, SEE-VCN is a straightforward MTDA add-on to bridge the scan pattern domain gap without requiring any architecture or loss modifications for any lidar-based detectors \cite{shi2020pv,shi2019pointrcnn,yan2018second,zhou2018voxelnet}.

\subsection{Point Cloud Completion}
Point cloud completion approaches focus on recovering complete point clouds from partial point clouds. Most approaches are proposed for synthetic 3D CAD objects \cite{chang2015shapenet,wu2015modelnet40} to allow evaluation with ground-truth object shapes. However, simulated sparse point clouds are not representative of the noise and sparsity of the point cloud of objects as captured from a real-world sensor. A few approaches \cite{yuan2018pcn,stutz2018learning,duggal2022mending} tackle completion for real-world lidar data, however, they rely on using ground truth bounding boxes to transform each object into object-centred coordinates. This access to ground truth annotations is infeasible in real-world applications. Conversely, there are few existing point cloud completion approaches for objects in viewer-centred coordinates. Viewer-centred coordinates represent the object in the coordinate frame aligned to the viewing perspective of the sensor. Traditional methods \cite{bernardini1999ball,edelsbrunner1983shape,kazhdan2006poisson} are typically able to complete viewer-centred objects, though the meshes are often inadequate for sparse observations, and require heavy parameter tuning for each object. Whilst deep learning models for object-centred completion have displayed great performance, adapting these models to the viewer-centred context is non-trivial. Gu et al. \cite{gu2020weakly} is the first method to estimate object pose and shape given an object in viewer-centred coordinates. Their training setup uses multiple views of the object in a Siamese training setup, with the ground truth surface as the accumulation of points from multiple views of that object. Our proposed VCN differs from this in that we present a synthetic-to-real approach, which is able to accommodate viewer-centred surface completion for various lidar scan patterns. We go a step further to demonstrate its application and upper-bound potential in the task of UDA for 3D object detection.

\section{Methodology}

SEE-VCN addresses the lidar scan pattern discrepancy through training and testing any 3D detector on a unified point cloud representation of objects. This is achieved through our viewer-centred surface completion network (VCN). For each object $\mathbf{P}_{vc}$, VCN transforms the object to a canonical frame $\mathbf{P}_{cn}$ before completing the surface. To adapt it for the UDA setting, we adopt the MTDA framework, SEE \cite{tsai2022see} and integrate it with VCN as shown in \cref{fig:framework}. In this paper, the object’s canonical frame (aka. object-centred coordinates) is defined as the coordinate frame where the object is centred at the origin, and aligned to the positive x-axis. Unlike ShapeNet \cite{chang2015shapenet} models, our canonical frame does not involve normalizing the objects to a unit-bounding box.

\subsection{Viewer-centered Surface Completion Network (VCN)}
\subsubsection{Pose Estimation}

To estimate the object pose, we follow \cite{qi2018frustum} by first transforming the object points $\mathbf{P}_{vc}$ to the frustum frame coordinates $\mathbf{P}_f$ to constrain the search space and improve rotation-invariance. The frustum angle is computed with $\theta_f = tan^{-1}(\frac{\mathbf{\bar{P_y}}}{\mathbf{\bar{P_x}}})$ where $\mathbf{\bar{P_x}}$ and $\mathbf{\bar{P_y}}$ are the mean x and y-axis coordinates of the object. Thereafter, we transform the input point cloud to the frustum frame $\mathbf{P}_f$ by applying rotation matrix $\mathbf{R}_y$ with angle $-\theta_f$ along the y-axis $\mathbf{P}_f = \mathbf{P}_{vc}\mathbf{R}_y (-\theta_f)$.

\begin{figure*}
  \vspace{2mm}
  \centering
  \includegraphics[width=0.99\linewidth]{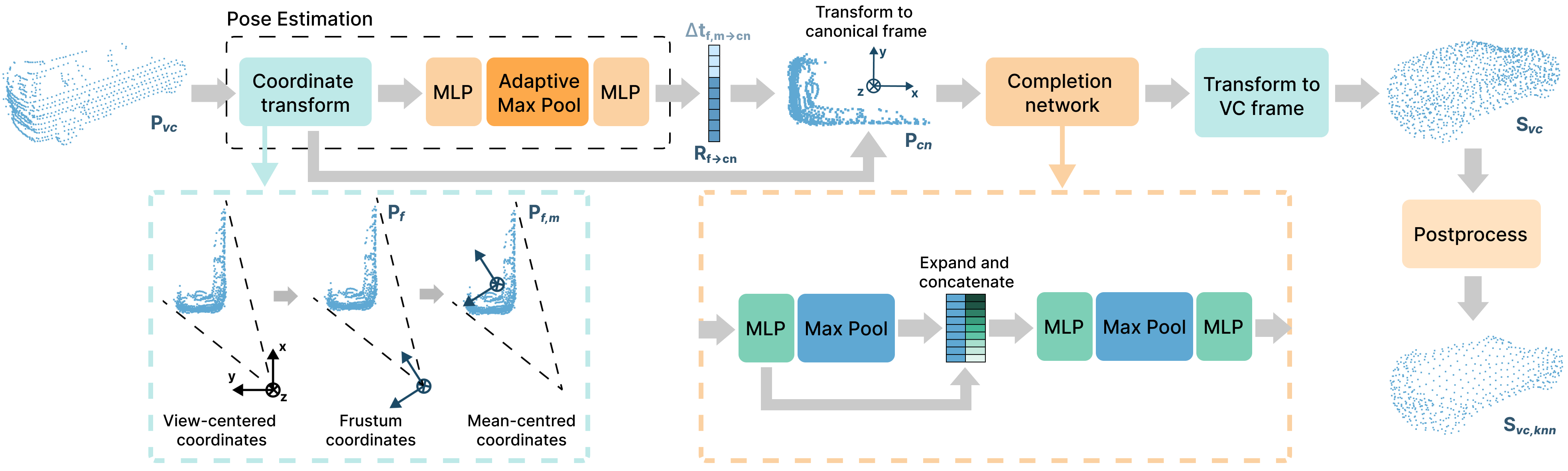}
  \caption{Architecture of our Viewer-centered surface Completion Network (VCN).}
  \label{fig:vcn_architecture}
  \vspace{-4mm}
\end{figure*}

We centre $\mathbf{P}_f$ on the mean of the points in the frustum frame with $\mathbf{P}_{f,m} = \mathbf{P}_f - \mathbf{\bar{P}}_f$ to get the frustum-mean frame coordinates $\mathbf{P}_{f,m}$ as illustrated in \cref{fig:vcn_architecture}. 

We feed the  $\mathbf{P}_{f,m}$ into our pose estimation component of VCN to get $ \mathbf{R}_{f\rightarrow cn},\mathbf{\Delta t}_{f,m\rightarrow cn} = \psi_{\text{pose}}({\mathbf{P}_{f,m}})$. VCN's pose estimation, $\psi_{\text{pose}}$, is comprised of 2 MLP blocks with Leaky ReLU activation, and an adaptive max pooling in between. $\psi_{\text{pose}}$ outputs a 9D vector which we split into a 3D translation delta $\mathbf{\Delta t}_{f\rightarrow cn}$, and a 6D rotation representation that we convert to a $3 \times 3$ matrix, $\mathbf{R}_{f\rightarrow cn}$, following \cite{zhou2019continuity} to ensure orthogonalization and rotation continuity. The point cloud in frustum frame $\mathbf{P}_{f}$ can now be transformed to a canonical frame with $\mathbf{P}_{cn} = (\mathbf{P}_{f\rightarrow cn} - \mathbf{t}_{f \rightarrow cn})\mathbf{R}_{f\rightarrow cn}$ where $\mathbf{t}_{f \rightarrow cn} = \mathbf{\bar{P}}_f + \mathbf{\Delta t}_{f,m\rightarrow cn}$ and $\mathbf{R}_f$ is the translation and rotation from frustum frame $\mathbf{P}_{f}$ to the canonical frame $\mathbf{P}_{cn}$ respectively. The pose from viewer-centred frame to canonical frame can be recovered by $\mathbf{R}_{vc\rightarrow cn} = \mathbf{R}_f \mathbf{R}_y(\theta_f)$ and $\mathbf{t}_{vc\rightarrow cn} = \mathbf{t}_f \mathbf{R}_y(\theta_f)$.

We use Geodesic loss for $\mathbf{R}_{vc\rightarrow cn}$, shown in \cref{eqn:geodesic1,eqn:geodesic2} with estimated rotation, $\mathbf{R}_{\text{pred}}$, and ground truth rotation, $\mathbf{R}_{\text{gt}}$. We use Smooth-$l_1$ (huber) loss for $\mathbf{t}_{vc\rightarrow cn}$. Combined, pose loss is $L_{\text{pose}} = \alpha L_{\mathbf{t}} + L_{\mathbf{R}}$ with weight $\alpha$.

\setlength{\belowdisplayskip}{6pt} \setlength{\belowdisplayshortskip}{6pt}
\setlength{\abovedisplayskip}{0pt} \setlength{\abovedisplayshortskip}{0pt}
\begin{equation}
\mathbf{R}'' = \mathbf{R}_{\text{pred}}\mathbf{R}_{\text{gt}}
\label{eqn:geodesic1}
\end{equation}
\begin{equation}
L_{rot} = \text{cos}^{-1}((\mathbf{R}''_{00} + \mathbf{R}''_{11} + \mathbf{R}''_{22} - 1)/2)
\label{eqn:geodesic2}
\end{equation}

\subsubsection{Surface Completion}
For surface completion, we adopt PCN's encoder \cite{yuan2018pcn}, which is a variant of PointNet \cite{qi2017pointnet}. Each PointNet module consists of an MLP module, followed by Max Pooling. PointNet uses an MLP for each point, $p_i \in \mathbf{P}_{cn}$, to transform $\mathbf{p}_i$ into a feature vector $\mathbf{f}_i$. These feature vectors are fed into a point-wise max pooling to obtain a global feature vector $\mathbf{g}$ that can intuitively be perceived as a summary of the information in $\mathbf{P}_{cn}$. For the output of the first PointNet, we concatenate the global vector $\mathbf{g}_1$ to each point feature vector $\mathbf{f}_i$ and feed it to a second PointNet module. The output of the second PointNet is our final global vector $\mathbf{g}_2$, which we feed into another MLP to reshape the global vector into the final object points $\mathbf{S}_{cn}$ of shape $N \times 3$. 

The output of the encoder network is commonly identified as ``coarse" or ``key" points in shape completion literature \cite{yuan2018pcn,yu2021pointr,qi2017pointnet}. $\mathbf{g}_2$ is commonly fed into a decoder network to generate a more fine-grained object shape. For our use case, the coarse form is sufficient as we aim to generate a general shape outline of cars with a unified representation across scan patterns. 

To obtain our completed car in viewer-centred coordinates, we first transform the object back to the frustum coordinates, then undo the frustum rotation as shown in \cref{eqn:completed_cn_to_vc}.

\setlength{\belowdisplayskip}{6pt} \setlength{\belowdisplayshortskip}{6pt}
\setlength{\abovedisplayskip}{-3pt} \setlength{\abovedisplayshortskip}{-3pt}
\begin{equation}
\mathbf{S}_{vc} = (\mathbf{S}_{cn}\mathbf{R}_f + \mathbf{t}_f ) \mathbf{R}_y(\theta_f)
\label{eqn:completed_cn_to_vc}
\end{equation}

In synthetic point cloud completion approaches, objects are typically pre-canonicalized into a unit-bounding box \cite{sun2021canonical}. However, in our scenario, we do not have the knowledge of object dimensions beforehand for pre-canonicalization. Therefore, to generate cars with more accurate sizes, we add a smooth-$l_1$ loss for the object dimension error $L_{\text{dim}}$ between the ground truth and predicted bounding box. When an object is in the canonical frame, it is trivial to estimate a 3D bounding box by taking the maximum and minimum bounds of the object points in each axis. 

To generate car-like surfaces, we follow existing works \cite{yuan2018pcn,yu2021pointr} to calculate the distance between each generated surface point $s \in \mathbf{S}_{vc}$, and the ground-truth surface point $g \in \mathbf{G}_{vc}$ using Chamfer Distance (CD) as our loss $L_{\text{complete}}$, shown in \cref{eqn:chamfer}. 

\setlength{\belowdisplayskip}{3pt} \setlength{\belowdisplayshortskip}{3pt}
\setlength{\abovedisplayskip}{-6pt} \setlength{\abovedisplayshortskip}{-6pt}

\begin{equation}\begin{split}
\text{CD $(\mathbf{S}_{vc},\mathbf{G}_{vc})$} = &\frac{1}{|\mathbf{S}_{vc}|}\sum_{s \in \mathbf{S}_{vc}} \min_{g \in \mathbf{G}_{vc}} ||s-g||_2 \\ 
+ &\frac{1}{|\mathbf{G}_{vc}|}\sum_{g \in \mathbf{G}_{vc}} \min_{s \in \mathbf{S}_{vc}} ||g-s||_2
\end{split}
\label{eqn:chamfer}
\end{equation}

\subsubsection{Post-Processing} 
Viewer-centred coordinates encodes the pose of the object. For certain partially observed objects, the generated 3D surface by VCN is good, but the pose is misaligned. This causes confusion for the 3D detector. For example, a common distribution of points on a car in KITTI \cite{geiger2012kitti} is where there are points on the rear, but none on the side, top or front. Given this set of points, it is difficult to determine a ground truth orientation; there could be multiple possible orientations. Dimensions are also challenging to estimate, since the rear of a sedan can appear similar to the rear of a limousine. We observe, however, that the surfaces generated by VCN are most accurate near the object's points $\mathbf{P}_{vc}$. Therefore, for each car, we retain the nearest N generated surface points  $s_i \in \mathbf{S}_{vc,knn}$ to each input object point $p_i \in \mathbf{P}_{vc}$. To improve the quality of $\mathbf{S}_{vc,knn}$, we add another CD loss $L_{\text{knn}}$ to minimise the distance between $\mathbf{S}_{vc,knn}$ and $\mathbf{G}_{vc,knn}$. 

We jointly optimise our network over the overall loss function shown in \cref{eqn:overall_loss}.

\begin{equation}
L = L_{\text{complete}} + L_{\text{knn}} + L_{\text{dims}} + L_{\text{pose}}
\label{eqn:overall_loss}
\end{equation}

\subsection{Training Data}
VCN was trained in a supervised manner, where partial $\mathbf{P}_{vc}$ and complete $\mathbf{G}_{vc}$ synthetic cars were prepared with ShapeNet \cite{chang2015shapenet} car models. To obtain the complete ground truth surface points $\mathbf{G}_{vc}$ for each car model, we applied raycasting from multiple views. The surface points from each viewpoint were concatenated together. 16384 points were thereafter sampled using farthest point sampling. To generate realistically positioned and occluded partial cars $\mathbf{P}_{vc}$, we used ``Vehicle" labels from the Waymo dataset to position and size our car models in a raycasting scene. Using the location of ``signs" from the Waymo labels, we placed poles of various heights. Once the scene was set up, we raycasted from the location of the lidar, to obtain partial cars in viewer-centred coordinates. 

\subsection{Simulated Lidar}
To ensure that VCN is not biased to a particular scan pattern, we simulate various sampling patterns for training. In the dataset generation, we densely raycast each partial car. This enables us to selectively downsample the partial cars to simulate different sampling patterns. We first convert each point into spherical coordinates $(r,\theta,\phi)$, and discretize vertical angle $\phi$ into N bins, each representing a lidar ring. For training, we randomly select every N\textsuperscript{th} ring, and every k\textsuperscript{th} point within the ring for each car as the input partial car.

\begin{table*}
\vspace{2mm}
\centering
\scalebox{0.95}{ 
\begin{tabular}{l|c|c|c|l|c} 
\toprule
Datasets & \multicolumn{1}{l|}{VFOV} & \multicolumn{1}{l|}{HFOV} & \multicolumn{1}{l|}{Beams}      & Sensors                                               & \multicolumn{1}{l}{Vertical Resolution (\textdegree)}  \\ 
\hline
KITTI    & 26.8                      & 360                       & 64                              & 1x Velodyne HDL64E, 2x cameras                        & 0.40                                        \\
nuScenes & 40                        & 360                       & 32                              & 1x Velodyne HDL32E, 6x cameras                        & $\sim$1.25                       \\
Waymo    & 20                        & 360                       & 64                              & 1x long-range lidar, 4x short-range lidar, 6x cameras & $\sim$0.31                       \\
Baraja   & 30                        & 120                       & \multicolumn{1}{l|}{Up to 1000} & 1x Baraja Spectrum-Scan™, 1x camera                    & 0.025 - 0.2                                 \\
\bottomrule
\end{tabular}}
\caption{Overview of the sensor setup across datasets.}
\label{tab:datasets}
\vspace{-2mm}
\end{table*}

\begin{table*}
\centering
\setlength{\tabcolsep}{0.3em} 
\begin{tabular}{c|c|cccc|ccc|ccc|ccc} 
\toprule
\multicolumn{1}{l}{} & \multicolumn{1}{l|}{} & \multicolumn{4}{c|}{VC-ShapeNet}                                                & \multicolumn{3}{c|}{KITTI}                                     & \multicolumn{3}{c|}{Waymo}                                     & \multicolumn{3}{c}{nuScenes}                                    \\ 
\hline
Model                & Config                & CD             & BEV / 3D             & R (\textdegree)            & t (m)          & BEV / 3D                      & R (\textdegree)            & t (m)          & BEV / 3D                     & R (\textdegree)            & t (m)          & BEV / 3D                      & R (\textdegree)            & t (m)           \\ 
\hline
PCN                  & -                     & 7296.31        & 0.02 / 0.01          & -             & -              & 0.02 / 0.01                   & -             & -              & 0.02 / 0.01                   & -             & -              & 0.02 / 0.01                   & -             & -               \\
VCN                  & -                     & 7575.57        & 0.04 / 0.03          & -             & -              & 0.04 / 0.03                   & -             & -              & 0.05 / 0.01                   & -             & -              & 0.06 / 0.05                   & -             & -               \\ 
\hline
VCN                  & c                     & 49.45          & 86.9 / 79.8          & -             & 0.083          & \textbf{81.6} / 72.5          & -             & 0.101          & 80.1 / 68.1                   & -             & 0.146          & 80.5 / 68.9                   & -             & 0.107           \\
VCN                  & c,r                   & 47.24          & 87.3 / 80.1          & \textbf{1.73} & 0.091          & 81.1 / 72.9                   & 2.96          & 0.108          & 80.4 / 68.9                   & \textbf{3.36} & 0.155          & \textbf{80.5} / 69.5          & \textbf{1.85} & 0.113           \\
VCN         & c,r,l                 & \textbf{39.12} & \textbf{88.1 / 81.5} & 1.77          & \textbf{0.080} & 80.6\textbf{ }/\textbf{ 74.3} & 3.30          & \textbf{0.099} & \textbf{80.9 }/\textbf{ 71.2} & 3.57          & \textbf{0.149} & 80.3\textbf{ }/\textbf{ 72.0} & 2.06          & \textbf{0.102}  \\
Atlas-VC \cite{groueix2018atlas}             & c,r,l                 & 56.17          & 83.4 / 76.1          & 2.48          & 0.094          & 78.4 / 69.8                   & 2.57          & 0.106          & 76.6 / 65.3                   & 4.60          & 0.164          & 77.5 / 68.1                   & 2.82          & 0.111           \\
SCiW \cite{gu2020weakly}               & c,r,l                 & 65.03          & 81.2 / 73.1          & 1.95          & 0.113          & 74.8 / 66.1                   & \textbf{2.31} & 0.125          & 75.3 / 63.2                   & 3.37          & 0.177          & 75.8 / 63.9                   & 2.44          & 0.134           \\
\bottomrule
\end{tabular}
\caption{Evaluation of viewer-centred surface completion of cars with at least 30 points. c: centre estimation, r: rotation estimation, l: lidar simulation used. R is median rotation error (\textdegree ) and t is mean translation error (m). BEV/3D refers to the IoU of the completed car's box compared with the ground truth box size. CD is evaluated with the L2 norm and scaled by 1000 (following \cite{yu2021pointr}).}
\label{tab:vcn_results}
\vspace{-6mm}
\end{table*}

\subsection{Augmented Point Clouds for 3D Object Detection}
For training and testing, we use two configurations of VCN: (1) VCN-CN is our object-centred configuration where we use ground truth labels to first canonicalize the car, before completing the surface. Pose estimation is not necessary for this setting. (2) VCN-VC is our viewer-centred configuration where the pose and surface of the car are both estimated. For training, we isolate the cars with the provided ground truth bounding boxes, and use VCN-CN to complete cars in the source domain. This augmented point cloud dataset is used for training the 3D detector. For testing, we follow \cite{tsai2022see} in isolating point cloud car instances $\mathbf{P}_{vc}$ by keeping the points within an image instance segmentation mask. These points are further clustered with DB-Scan \cite{ester1996density} before passing to VCN-VC for surface completion of each car. Each car instance $\mathbf{P}_{vc}$ is replaced with the completed, and post-processed car $\mathbf{S}_{vc,knn}$ to form our final augmented point cloud for inference.

\section{Experiments}
We evaluate SEE-VCN on the ``car"/``vehicle" category across multiple datasets, summarised in \cref{tab:datasets}. We assess the following scenarios: (1) Same number of beams (Waymo $\rightarrow$ KITTI). Despite having the same number of beams, there still remains a performance degradation. This is in part due to object size \cite{wang2020train}, but also from scan pattern variations \cite{tsai2022see}; (2) Low $\rightarrow$ high beam (nuScenes $\rightarrow$ KITTI). We show that we can attain good performance by upsampling objects of interest instead of the entire point cloud \cite{li2019pu,yifan2019patch,yi2021complete}. (3) High $\rightarrow$ low beam. (4) Lidar with adjustable scan pattern (Waymo/nuScenes $\rightarrow$ Baraja Spectrum-Scan™).

\subsection{Viewer-Centred Surface Completion}
We first evaluate the completion ability of VCN for cars from both synthetic and real datasets in this section. 

\noindent
{\bf VC-ShapeNet.} Our generated viewer-centred data for VCN consists of 3394 car models, split into 3054 for training and 340 for testing. For the training set, we have 20 different partial views per car model, giving 61,080 total training samples. For the testing set, we have a total of 5100 samples (15 partial views per car), sampled with our simulated lidar. 

\noindent
{\bf Lidar Data.} We form a nuScenes, KITTI and Waymo test dataset, each consisting of 5000 cars with at least 30 points. 

\noindent
{\bf Evaluation.} To evaluate the completed car, we use two metrics. (1) 3D and BEV IoU of the completed car with the ground truth car bounding box for both synthetic and real datasets; (2) CD with the ground truth car surfaces of the synthetic dataset. For comparison, we pick two methods, SCiW and Atlas-VC. SCiW \cite{gu2020weakly} performs single-frame surface completion for viewer-centred objects using a Siamese training setup. As they did not release their code, we therefore compare with our implemented adaptation of it. We adapted SCiW to train on VC-ShapeNet instead of real data for fair comparison. Atlas-VC is based on AtlasNet \cite{groueix2018atlas}. We use VCN's pose estimation to estimate pose, followed by AtlasNet's patch-based approach to complete the surface. 

\noindent
{\bf Results.} \cref{tab:vcn_results} shows that a direct implementation of object-centred networks such as PCN \cite{yuan2018pcn} fails to generate meaningful complete shapes. We show that VCN without pose estimation also suffers similarly. Simply estimating the centre allows VCN to begin generating meaningful surfaces and more accurate car dimensions. By adding a rotation estimation, we can further decrease the CD and increase IoU. When we simulate the lidar scan patterns, we further boost each real lidar by 2-3\% in 3D IoU box estimation, as well as decrease CD. VCN outperforms Atlas-VC and SCiW in both CD and IoU metrics. We also evaluate the median rotation and mean translation error and show that for object pose estimation, VCN can obtain a median rotation error of 2-3\textdegree \space and a translation error of 0.10-0.14m. 

\begin{table*}
\vspace{2mm}
\centering
\scalebox{0.95}{ 
\begin{tabular}{c|c|c|c|c|c} 
\toprule
\multirow{2}{*}{Task}             & \multirow{2}{*}{Method} & \multicolumn{2}{c|}{SECOND-IoU}                   & \multicolumn{2}{c}{PV-RCNN}                        \\ 
\cline{3-6}
                                  &                         & $\text{AP}_\text{BEV}$ / $\text{AP}_\text{3D}$                        & Improvement     & $\text{AP}_\text{BEV}$ / $\text{AP}_\text{3D}$                        & Improvement      \\ 
\hline
\multirow{7}{*}{Waymo $\rightarrow$ KITTI}    & Source-only             & 71.82 / 38.33                   & -               & 69.06 / 39.97                   & -                \\ 
\cline{2-6}
                                  & ST3D                    & 78.93 / 62.81                   & +07.11 / +24.48 & 89.94 / 73.32                   & +20.88 / +33.35  \\
                                  & SEE                     & \textbf{88.09} / 65.52          & +16.27 / +27.19 & \textbf{90.82} / \textbf{79.39} & +21.76 / +39.42  \\
                                  & SEE-VCN                   & 87.19 / \textbf{69.27}          & +15.37 / +30.94 & 89.00 / 74.68                   & +19.94 / +34.71  \\ 
\cline{2-6}
                                  & SEE (Ideal iso)         & 88.93 / 70.48                   & +17.11 / +32.15 & 93.17 / 84.49                   & +24.11 / +44.52  \\
                                  & SEE-VCN (Ideal iso)       & 93.25 / 79.02                   & +21.43 / +40.69 & 94.08 / 85.84                   & +25.02 / +45.87  \\ 
\cline{2-6}
                                  & Oracle                  & 86.07 / 81.73                   & -               & 92.17 / 89.32                   & -                \\ 
\hline\hline
\multirow{7}{*}{Waymo $\rightarrow$ nuScenes} & Source-only             & 81.54 / 57.99                   & -               & 80.23 / 65.26                   & -                \\ 
\cline{2-6}
                                  & ST3D                    & \textbf{85.85 / 66.71}          & +04.31 / +08.72 & \textbf{85.59 / 71.93}          & +05.36 / +06.67  \\
                                  & SEE                     & 74.37 / 44.93                   & -07.17 / -13.06 & 75.63 / 55.05                   & -04.60 / -10.21  \\
                                  & SEE-VCN                   & 69.03 / 39.28                   & -12.51 / -18.71 & 76.51 / 50.98                   & -03.72 / -14.28  \\ 
\cline{2-6}
                                  & SEE (Ideal iso)         & 82.47 / 53.80                   & 00.93 / -04.19  & 85.60 / 67.38                   & 05.37 / 02.12    \\
                                  & SEE-VCN (Ideal iso)       & 82.57 / 56.30                   & +01.03 / -01.69 & 87.33 / 52.45                   & +07.10 / -12.81  \\ 
\cline{2-6}
                                  & Oracle                  & 86.05 / 71.92                   & -               & 85.03 / 70.94                   & -                \\ 
\hline\hline
\multirow{7}{*}{nuScenes $\rightarrow$ KITTI} & Source-only             & 47.20 / 10.87                   & -               & 64.66 / 43.27                   & -                \\ 
\cline{2-6}
                                  & ST3D                    & 76.82 / 58.58                   & +29.62 / +47.71 & \textbf{88.89} / \textbf{83.86} & +24.23 / +40.59  \\
                                  & SEE                     & 77.00 / 56.00                   & +29.80 / +45.13 & 85.53 / 72.51                   & +20.87 / +29.24  \\
                                  & SEE-VCN                   & \textbf{85.23} / \textbf{71.02} & +38.03 / +60.15 & 86.88 / 76.01                   & +22.22 / +32.74  \\ 
\cline{2-6}
                                  & SEE (Ideal iso)         & 83.80 / 66.82                   & +36.60 / +55.95 & 89.94 / 80.29                   & +25.28 / +37.02  \\
                                  & SEE-VCN (Ideal iso)       & 94.74 / 83.54                   & +47.54 / +72.67 & 92.78 / 85.92                   & +28.12 / +42.65  \\ 
\cline{2-6}
                                  & Oracle                  & 86.07 / 81.73                   & -               & 92.17 / 89.32                   & -                \\
\bottomrule
\end{tabular}}
\caption{Results for the different DA scenarios, reported with an IoU threshold of 0.7. ``Improvement" denotes the AP increase from source-only to the various approaches in the format $\text{AP}_\text{BEV}$ / $\text{AP}_\text{3D}$. "Ideal iso" refers to the scenario where car points are ideally isolated in testing. All models are evaluated on ground truth boxes with 50 points or more.}
\label{tab:main_results}
\vspace{-3mm}
\end{table*}

\subsection{Unsupervised Domain Adaptation}
In this section, we evaluate SEE-VCN for UDA on two 3D detectors, SECOND-IoU \cite{yan2018second} and PV-RCNN \cite{shi2020pv}. For training of Waymo and nuScenes, we subsample the data to use 5267 frames for Waymo and 4025 frames for nuScenes. 

\noindent
{\bf Baraja Spectrum-Scan™.} We build upon the dataset collected by \cite{tsai2022see} to increase the number of frames and scan pattern variations. The dataset has 408 frames, with 3 different types of scan configurations: (1) 64-beam, dense at the centre of the vertical FOV and sparse at the extremities; (2) 128-beam uniformly distributed across the vertical FOV; (3) Horizon-tracking (96-beam), where the scan pattern is adjusted whilst driving to have higher density at the horizon.

\noindent
{\bf Evaluation.} We evaluate results on Average Precision (AP) at the 0.7 threshold for 3D and BEV IoU. For KITTI, we report the moderate AP. We compare SEE-VCN with (1) Source only, where no DA is used; (2) ST3D \cite{yang2021st3d}, the SOTA DA method for single-frame UDA; (3) SEE \cite{tsai2022see}, the only existing MTDA method; (4) Oracle, the model trained on the target domain. We additionally include the ideal isolation scenario (``Ideal iso") where we use ground truth boxes to isolate the car points for input to VCN-VC. This shows the upper bound of SEE-VCN if we improve on the isolation of points for inference. 

\noindent
{\bf Results.} \cref{tab:main_results} shows that SEE-VCN outperforms ST3D and SEE in Waymo $\rightarrow$ KITTI and nuScenes $\rightarrow$ KITTI for SECOND-IoU. For PV-RCNN in nuScenes $\rightarrow$ KITTI, SEE-VCN outperforms SEE, but underperforms in  Waymo $\rightarrow$ KITTI. However, when assessing the ideal isolation scenario, we observe that when VCN is fed with the correctly isolated points, SEE-VCN is able to achieve a significant 10-17\% increase over SEE in 3D AP. This improvement highlights the potential of SEE-VCN when the isolation of points is improved. In the Waymo $\rightarrow$ nuscenes scenario, SEE-VCN suffers due to the low resolution in nuScenes dataset. Due to the low resolution, any noise would exacerbate the errors of VCN. In the future, we expect this to be less of an issue as lidar manufacturing is headed towards higher point cloud resolution. Overall, SEE-VCN performs comparatively to ST3D, with the advantage of being an MTDA approach. We demonstrate this on the Baraja dataset without re-training in \cref{tab:baraja_results}, where SEE-VCN obtains higher performance than source-only and SEE in multiple DA settings. If we can obtain perfect isolation of points (ideal iso), SEE-VCN can further increase the performance by around 12-25\% AP, which is 18-33\% better than SEE (ideal iso). 

\begin{table}
\centering
\scalebox{0.94}{ 
\begin{tabular}{c|c|c|c} 
\toprule
\multirow{2}{*}{Source}   & \multirow{2}{*}{Method} & SECOND-IoU    & PV-RCNN        \\ 
\cline{3-4}
                          &                         & $\text{AP}_\text{BEV}$ / $\text{AP}_\text{3D}$      & $\text{AP}_\text{BEV}$ / $\text{AP}_\text{3D}$       \\ 
\hline
\multirow{5}{*}{Waymo}    & Source-only             & 83.52 / 71.98 & 83.27 / 76.76  \\ 
\cline{2-4}
                          & SEE                     & \textbf{87.39 / 77.10} & 82.77 / 74.60  \\
                          & SEE-VCN                   & 86.10 / 75.12 & \textbf{88.34 / 80.50}  \\ 
\cline{2-4}
                          & SEE (Ideal iso)         & 87.09 / 70.99 & 85.52 / 79.75  \\
                          & SEE-VCN (Ideal iso)       & 94.27 / 88.93 & 93.66 / 90.53  \\ 
\hline\hline
\multirow{5}{*}{nuScenes} & Source-only             & 16.88 / 12.00 & 18.31 / 15.30  \\ 
\cline{2-4}
                          & SEE                     & 52.23 / 34.63 & 71.19 / 62.15  \\
                          & SEE-VCN                   & \textbf{66.63 / 47.45} & \textbf{78.04 / 63.75}  \\ 
\cline{2-4}
                          & SEE (Ideal iso)         & 52.97 / 36.73 & 72.36 / 65.54  \\
                          & SEE-VCN (Ideal iso)       & 85.59 / 70.86 & 90.16 / 82.14  \\
\bottomrule
\end{tabular}}
\caption{We demonstrate that SEE-VCN on the Baraja Spectrum-Scan™ dataset with an IoU threshold of 0.7.}
\label{tab:baraja_results}
\vspace{-3mm}
\end{table}

\begin{table}
\centering
\begin{tabular}{l|llll} 
\toprule
         & \multicolumn{4}{c}{$\text{IoU}_\text{BEV}$ / $\text{IoU}_\text{3D}$}                                                 \\ 
\cline{2-5}
         & \textgreater{} 201 & 81 - 200 & 31 - 80 & 5 - 30  \\ 
\hline
KITTI    & 84.0 / 78.5                 & 79.8 / 73.9       & 76.1 / 67.9      & 62.8 / 49.7      \\
nuScenes & 81.3 / 75.6                 & 81.2 / 73.9       & 78.7 / 67.8      & 65.2 / 50.3      \\
Waymo    & 85.5 / 77.8                 & 80.6 / 70.4       & 75.9 / 64.0      & 58.8 / 46.1      \\
\bottomrule
\end{tabular}
\caption{IoU of VCN for cars with varying number of points on lidar datasets.}
\label{tab:vcn_varying_pts}
\vspace{-3mm}
\end{table}

\subsection{Ablation}

\noindent
{\bf VCN at varying number of points.} \cref{tab:vcn_varying_pts} shows that when there is above 200 points, VCN's completion performs well across datasets. When there is less than 30 points however, the performances degrades significantly. 

\noindent
{\bf Post-processing.} In \cref{tab:postproc}, we show that directly using VCN causes more sensitivity to the errors in surface completion, leading to lower detector performance. When we retain the points nearest to the input points (KNN), we are able to increase the performance by 6.34\% in 3D IoU. In certain scenarios, car points can be far apart, due to occlusion or noise. Therefore, to reduce the influence of noise, we further cluster $\mathbf{S}_{vc,knn}$ to retain only the generated surface points that are close to each other (KNN+DB).

\noindent
{\bf Minimum number of points.} We provide results for SEE-VCN for evaluation at various number of minimum points in \cref{tab:min_pts_table} for comparison with existing and future methods. 

\noindent
{\bf Synthetic vs real.} We observe in \cref{tab:vcn_results} that on synthetic data, VCN can obtain a higher 3D IoU and lower pose error compared to real data. Whilst the simulated lidar has alleviated the difference between the synthetic and real data, further work is required to close the gap.

\begin{table}
\centering
\begin{tabular}{l|l} 
\toprule
        & $\text{AP}_\text{BEV}$ / $\text{AP}_\text{3D}$  \\ 
\hline
KNN + DB-Scan & 87.19 / 69.27      \\
KNN     & 85.54 / 68.81      \\
No post-processing    & 81.16 / 62.47      \\
\bottomrule
\end{tabular}
\caption{IoU results for our post-processing step using SECOND-IoU on the Waymo $\rightarrow$ KITTI setting.}
\label{tab:postproc}
\vspace{-6mm}
\end{table}

\begin{table}
\centering
\setlength{\tabcolsep}{0.3em} 
\begin{tabular}{c|c|c|c|c} 
\toprule
\multicolumn{1}{l|}{} & \multicolumn{2}{c|}{Waymo $ \rightarrow $ KITTI} & \multicolumn{2}{c}{nuScenes $ \rightarrow $ KITTI}  \\ 
\hline
Min pts.              & SECOND-IoU     & PVRCNN             & SECOND-IoU    & PV-RCNN               \\ 
\hline
0                     & 73.35 / 56.31 & 75.22 / 61.38      & 72.79 / 58.47 & 76.98 / 64.01         \\
30                    & 82.96 / 63.40 & 84.77 / 69.13      & 81.92 / 65.90 & 84.33 / 71.61         \\
50                    & 87.19 / 69.27 & 89.00 / 74.68      & 85.23 / 71.02 & 86.88 / 76.01         \\
\bottomrule
\end{tabular}
\caption{Evaluation of SEE-VCN on ground truth boxes with 0, 30 and 50 points or more.}
\label{tab:min_pts_table}
\vspace{-6mm}
\end{table}

\subsection{Discussion}
The run time of the surface completion of SEE-VCN is 0.32ms (3125 cars/s) for VCN on a 2080Ti, and 20.1ms (50 cars/s) for post-processing on an Intel Core™ i7-6700K CPU. We observe that there is around 10\% difference in 3D IoU between SEE-VCN and SEE-VCN (Ideal iso), indicating that incorrectly isolated object points can negatively affect the completions of VCN. This could be due to misaligned calibration \cite{tsai2021optimising}, occlusions or incorrect/missed image detections. For instance, the image detector would occasionally include a trailer as part of a car, leading to ill-formed completed cars. In other cases, bushes, or poles may be very close to the car, causing it to be clustered as one object. Regardless, overcoming these challenges in object isolation for SEE-VCN can lead to better performance as shown in the ideal isolation upper bound. 

\section{Conclusion}
In this work, we propose SEE-VCN to address the scan pattern domain gap in point clouds acquired by different lidars in the task of UDA for 3D detection. We demonstrate that unifying the representation of the objects allows for more accurate detections. To our knowledge, this paper is the first to propose a synthetic-to-real and lidar-agnostic, viewer-centred surface completion network to estimate pose and generate object surfaces. We take it further to demonstrate its application in UDA for 3D detection with our MTDA approach, SEE-VCN. Through extensive experiments we show that SEE-VCN outperforms the previous MTDA SOTA, SEE in most DA settings. We can also outperform previous STDA work, ST3D, on multiple settings without needing to tune for a specific target domain. 

\bibliography{references}
\bibliographystyle{IEEEtran}

\end{document}